# Boosting Multimodal Remote Sensing Image Classification with Transformer-based Heterogeneously Salient Graph Representation

Jiaqi Yang, Bo Du, *Senior Member, IEEE,* Rong Liu, *Member, IEEE,* Zhu Mao, Liangpei Zhang, *Fellow, IEEE*

***Abstract*—Data collected by different modalities can provide a wealth of complementary information, such as hyperspectral image (HSI) to offer rich spectral-spatial properties, synthetic aperture radar (SAR) to provide structural information about the Earth's surface, and light detection and ranging (LiDAR) to cover altitude information about ground elevation. Therefore, a natural idea is to combine multimodal images for refined and accurate land-cover interpretation. Although many efforts have been attempted to achieve multi-source remote sensing image classification, there are still three issues as follows: 1) indiscriminate feature representation without sufficiently considering modal heterogeneity, 2) abundant features and complex computations associated with modeling long-range dependencies, and 3) overfitting phenomenon caused by sparsely labeled samples. To overcome the above barriers, a transformer-based heterogeneously salient graph representation (THSGR) approach is proposed in this paper. First, a multimodal heterogeneous graph encoder is presented to encode distinctively non-Euclidean structural features from heterogeneous data. Then, a self-attention-free multi-convolutional modulator is designed for effective and efficient long-term dependency modeling. Finally, a mean forward strategy is developed in order to avoid overfitting. Based on the above structures, the proposed model is able to break through modal gaps to obtain differentiated graph representation with competitive time cost, even for a small fraction of training samples. Experiments and analyses in three benchmark datasets with various state-of-the-art (SOTA) approaches show the performance of the proposed THSGR. The code will be available in https://github.com/jqyang22.**

## I. Introduction

WITH the improvement of data acquisition capabilities, it has become possible to obtain large amounts of publicly available remote sensing (RS) data from different sensors. Multimodal data enables various types of Earth observation missions, such as land-cover identification [1], urban studies [2], precision agriculture [3], and resource management [4], which contribute to the exploration and discovery of our living environment [5]. Among various multimodal data, hyperspectral image (HSI) encompasses hundreds of almost continuous spectral bands, allowing for detailed land-cover classification [6, 7]. Nevertheless, the recognition capability of HSI is restricted when faced with ground objects that have highly similar spectral signatures or spatial textures, such as the "Bricks" on the roof or on the path in complex urban scenarios. Fortunately, the structural properties from synthetic aperture radar (SAR) data or elevation information in light detection and ranging (LiDAR) imagery can assist in well distinguishing them with higher accuracy. Further, the passive imaging mechanism of HSI is easily interrupted by severe cloudy weather, resulting in intermittent imaging time and information loss. In contrast, SAR data has the all-weather active acquisition capability, while LiDAR provides active elevation measurements under clear-sky conditions. They can be considered as the auxiliary data to alleviate the above issues in HSI. Hence, combining HSI with SAR or LiDAR data for collaborative classification becomes a viable solution and has aroused wide attention. However, due to the differences in imaging sensors, context and resolution across modalities, how to integrate their complementary properties into an effective, compact and differentiated representation remains a challenging problem [8-11].

Many attempts have been made to realize multimodal land cover mapping [12, 13]. In the past few years, feature-level fusion studies were usually concatenated for classification and have been widely proposed for the multimodal classification task. In [14], morphological operators were first introduced to extract joint features of varying modalities. Then, multisource attributes were presented in [15] for invariant feature representation. Similar research included [16] and [17]. Unlike feature fusion-based works, decision fusion-based approaches produced the final results by aggregating decision-level

[1] This work was supported in part by the National Natural Science Foundation of China under Grant 62225113, in part by the Innovative Research Group Project of Hubei Province under Grant 2024AFA017, and in part by New Cornerstone Science Foundation through the XPLORER PRIZE. (*Corresponding author: Bo Du.*)

J. Yang is with Department of Forest and Wildlife Ecology, University of Wisconsin-Madison, Madison, WI 53706, USA (e-mail: jiaqi.yang@wisc.edu).

B. Du is with the School of Computer Science, National Engineering Research Center for Multimedia Software, Institute of Artificial Intelligence, and Hubei Key Laboratory of Multimedia and Network Communication Engineering, Wuhan University, Wuhan, China (e-mail: dubo@whu.edu.cn).

R. Liu is with the School of Geography and Planning, Sun Yat-sen University, Guangzhou 510275, China (e-mail: liurong25@mail.sysu.edu.cn).

Z. Mao is with the Department of Forest Sciences, University of Helsinki, Helsinki, 00014, Finland (e-mail: zhu.mao@helsinki.fi).

L. Zhang is with the State Key Laboratory of Information Engineering in Surveying, Mapping, and Remote Sensing, Wuhan University, Wuhan 430079, China (e-mail: zlp62@whu.edu.cn).

classification maps [18]. Liao et al. [19] presented a graph-based fusion method with weighted majority voting. Luo et al. [20] developed extended multi-attribute profiles for reliable classification. [21] was also a typical method. In summary, the above classical models rely heavily on experts' domain knowledge and manual tuning of parameters, which are difficult to be applied adaptively and performed quickly for classification tasks in different scenes [22]. Gao et al. [23] proposed a prototype-based information compensation network that focuses on inter-frequency communication. In [24], a spatial-spectral masked autoencoder was designed to extract local and high-frequency information simultaneously.

To automatically extract abstract and robust features from multiple data sources, the deep learning technique has been introduced into the field of heterogeneous remote sensing image classification [25]. Owing to the superiority in modeling hierarchical feature expression, convolutional neural network (CNN) has become a popular paradigm. Wang et al. [26] proposed a multi-attentive hierarchical fusion net to fuse the oriented features. Both fine- and coarse-grained information could be obtained from this CNN-based net, yet diagnostic features were insufficient to achieve fine classification. In [27], a multi-branch feature fusion architecture with self- and cross-guided attention was designed for exploiting hierarchical features in HSI and LiDAR. Although neighbor information from two modes was fused, it was accompanied by a large number of redundant features and training parameters. Li et al. [28] merged pixel-level multisource features via a novel spatial-spectral salient reinforcement network. Spectral-spatial-elevation properties were simultaneously captured, whereas overfitting problem caused by small-size labeled samples still occurred. A cross-channel reconstruction approach was presented in [29] for HSI-LiDAR and HSI-SAR union classification. Features captured by different data sources could be simultaneously constructed, but it came with a serious performance bottleneck due to the lack of diagnostic information. It can be noted that although CNN-based approaches yield considerable performance, they still suffer from insufficient variance characteristics, excessively useless information and parameters, as well as an overfitting phenomenon [30, 31].

Recently, the transformer has been successfully applied in many tasks owing to its strong ability to capture long-term dependencies, such as object detection, semantic segmentation, person re-identification, and video processing [31-33]. In the field of multimodal joint classification, some studies also emerged for improving the classification effect. A dual-branch network was utilized in [34] for learning spectral–spatial-elevation properties from LiDAR data. Multisource features could be gained on the net, notwithstanding missing saliently discriminatory information. As a result, the performance gain was unsatisfactory. Xue et al. [35] proposed a deep hierarchical vision transformer with a cross-attention feature fusion pattern to dynamically combine heterogeneous features. Under this circumstance, multiple attention mechanisms brought in more calculations and parameters. In [36], an attention fusion of transformer-based and scale-based method was created in order to encode spectral-spatial-elevation features. In this framework, heterogeneously structural information was reconstructed but still faced an overfitting issue. In fact, the above-mentioned transformer-based studies have tried to improve the accuracy, yet still struggle to overcome the following obstacles: undifferentiated heterogeneity feature extraction, complex models and redundant computational cost caused by non-local learning, and overfitting with small samples [37, 38].

Since graph convolutional network (GCN) is competent in finding relationships of different structures by prediction of nodes, edges, or graphs, several GCN-based works have been devised in the multimodal classification missions [39]. Xiu et al. [40] employed a multisource attention network with discriminative graphs and informative entities for multi-sensor information perception. Still, the net simply blended information from multiple sources rather than extracting heterogeneous and distinguishing features. In [41], a graph fusion network was designed to capture long-range information, while inevitably introducing redundant features that make little contribution to the classification. Guo et al. [42] attempted to bridge the data heterogeneity gap via dual graph convolution joint dense networks. Due to the scarcity of labeled data, the model lacked effective processing for avoiding overfitting and encountered serious performance degradation. Although these GCN-based approaches focus on mining heterogeneous structural attributes, they still retain the difficulties of salient feature extraction, feature and parameter redundancy, and accuracy reduction arising from overfitting [43, 44].

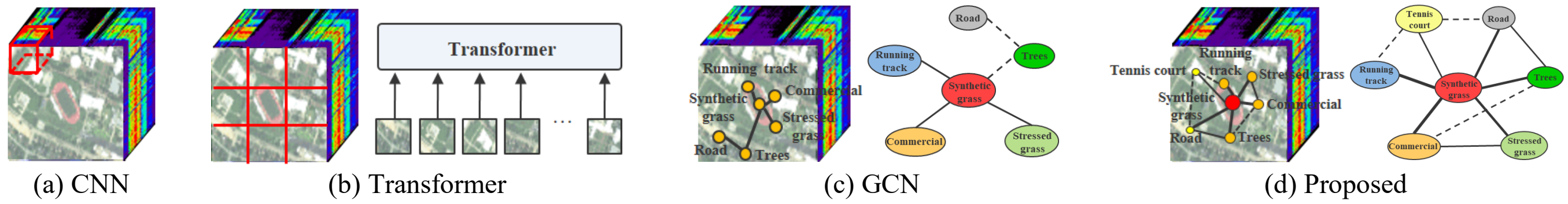

Fig. 1. The comparison of existing popular approaches with the proposed method (Thick lines, thin lines, and dashed lines in turn indicate the weakened structural relationship).

In view of the above analysis, a transformer-based heterogeneously salient graph representation (THSGR) approach is proposed to realize discriminative feature extraction of multimodal data, long-distance dependency modeling without redundant features or parameters as well as avoiding overfitting in this paper. Fig. 1 gives a visible comparison of the existing popular methods with the proposed model. It can be seen that CNN-based approaches usually take a patch in the neighborhood around a central pixel to extract features of spatial-channel dimension. Hence, global semantics and salient features are inadequate for high-precision cooperative classification. Transformer-related research always crops the input patch into multiple small blocks and then sends them to the transformer for multi-head self-attention (MSA) computation so as to obtain long-term dependencies. Whereas transformer-related approaches fail to capture local details and position information, resulting in undesirable performance. Actually, there are a number of useless features and calculations in MSA, that interferes with the learning of different modalities. GCN-relevant studies are able to learn structural relationships among various land covers, which is beneficial for non-local feature extraction. Nevertheless, existing GCN-based works tend to select fewer categories in close proximity to form the vertices, ignoring auxiliary categories that are geographically distant but significant for classification (e.g., “Tennis court” outside the “Synthetic grass” tends to appear together. If “Tennis court” are always neglected, the identification of “Synthetic grass” will be hampered). Moreover, these methods indiscriminately deal with the relationship between the category to be classified and other categories. Hence, it is challenging to extract heterogeneously salient features. Additionally, the above models are all easily over-fitted to the input due to the lack of labeled samples. By contrast, the proposed approach takes more categories into account and acquires more content-aware structural relationships. Specifically, structural relationships between the category to be classified and other categories, as well as those among other categories, are hierarchical. For example, “Synthetic grass” and “Running track” are closely related, “Synthetic grass” and “Road” are sub-associated, while “Commercial” and “Trees” are weakly correlated. To this end, salient graph representation can be generated for better classification performance. The detailed flowchart of the proposed method is shown in Fig. 2. The main contributions of this study are summarized as follows:

1) A multimodal heterogeneous graph encoder is first designed for multi-sensor data. In this way, the variably salient graph, rather than a single static adjacency matrix or globally shared representation, can be explicitly generated to achieve heterogeneously distinctive topological feature extraction.

2) Afterwards, a multi-convolutional modulator is presented to model long-term dependencies and to re-weight spectral-spatial patterns with fewer calculations and parameters. With a self-attention-free architecture, high-level semantics that are decisive and recalibrated can be better captured.

3) Additionally, we define a mean forward approach to mitigate overfitting. Leveraging a simple yet effective mechanism, this module facilitates an implicit ensemble that stabilizes predictions, particularly when training with limited samples.

4) Experimental results and analyses in three real-world HSI-SAR/LiDAR datasets demonstrate that the proposed approach can outperform other advanced approaches and yield superior classification performance.

The remainder of this paper is organized as follows. Section II presents the proposed method in detail. Section III reports experiments and analysis. Conclusion and future work are provided in Section IV.

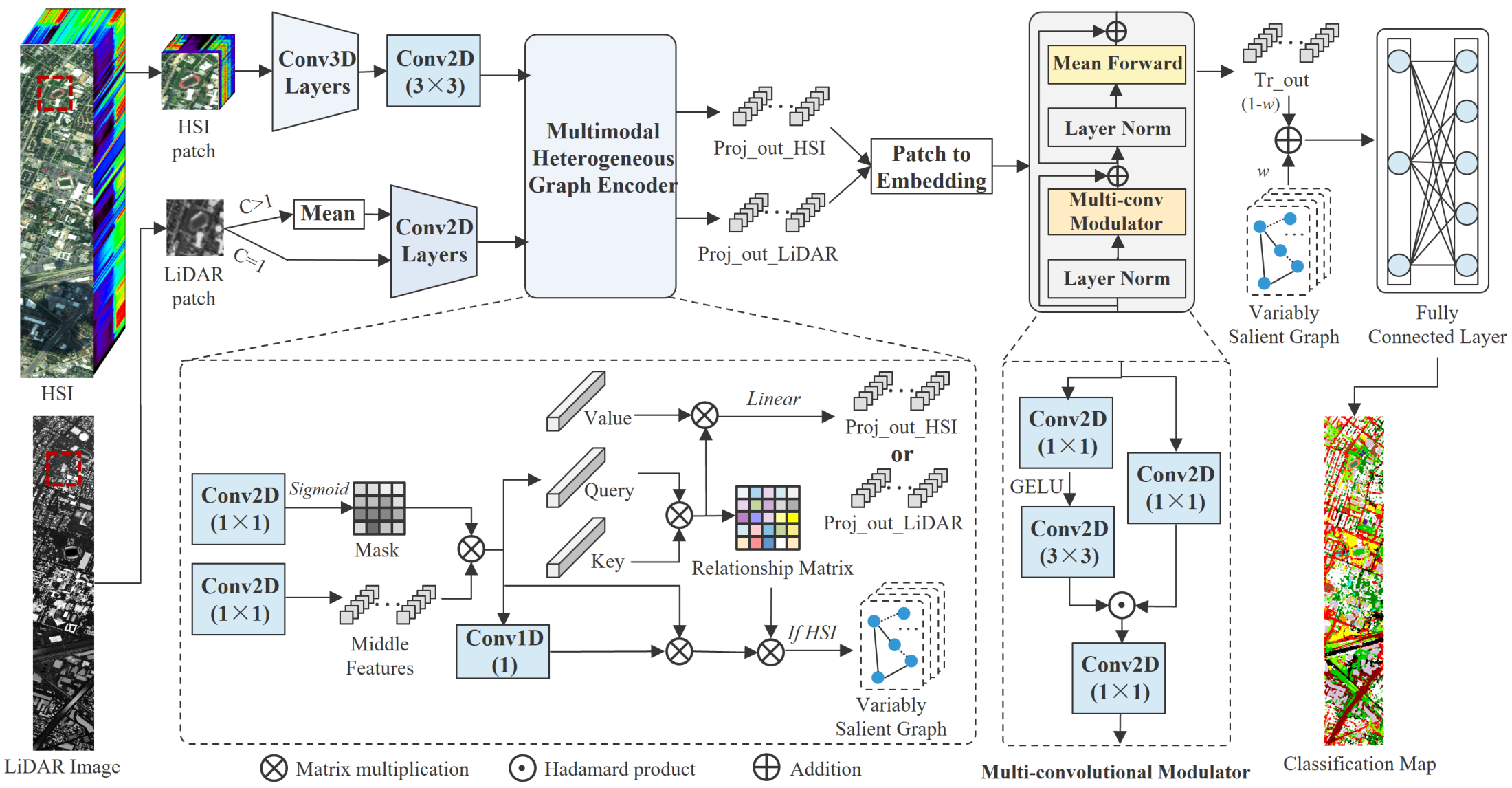

Fig. 2. The flowchart of the proposed THSGR. For each pixel, an HSI patch and the corresponding LiDAR patch are first extracted. The HSI patch is encoded by several 3-D and 2-D convolution layers to obtain spectral–spatial features, while the LiDAR patch is aggregated (optional mean operation) and encoded by 2-D convolutions. These modality-specific features are then fed into the **multimodal heterogeneous graph encoder** to construct a variably salient graph and two graph-enhanced projections. Two graph-enhanced projections are subsequently converted into token embeddings via patch-to-embedding and refined by the **multi-convolutional modulator**. A **mean-forward** operation is applied to aggregate the refined tokens into a compact representation, which is then fused with the variably salient graph and sent to a fully connected layer to predict the category label for each pixel, forming the classification map.

## II. Methodology

In this section, a detailed description of the proposed THSGR will be given, which includes a multimodal heterogeneous graph encoder, multi-convolutional modulator, and mean forward.

### *A. Multimodal Heterogeneous Graph Encoder*

Multi-source data can provide diverse structural information, but how to learn discriminative features and complementary information from them remains a challenge [45]. The proposed multimodal heterogeneous graph encoder illustrated in Fig. 2 offers a new viewpoint to address the above issue.

Let the input HSI and SAR/LiDAR imagery after regularization [46] be $X_{hsi} \in \mathbb{R}^{H\times W\times C_{hsi}}$ and $X_{s/l} \in \mathbb{R}^{H\times W\times C_{s/l}}$, where $H$, $W$, $C_{hsi}$ and $C_{s/l}$ denote the height, width, number of bands in the original HSI, and number of bands in the original SAR/LiDAR data, respectively. To mitigate Hughes phenomenon of high-dimensional HSI, PCA is employed for dimensionality reduction [47]. The dimension-reduced HSI can be derived as $X_{hsi}^{p} \in \mathbb{R}^{H\times W\times C_{hsi}^{p}}$, where $C_{hsi}^{p}$ represent the first retained $p$ principal components (PCs). Around each sample to be classified in $X_{hsi}^{p}$, extract a neighbor patch $x_{hsi}^{p}$ with the size of $k \times k \times C_{hsi}^{p}$, where $k$ is the spatial size of the patch. Similarly, $x_{s/l}$ in $X_{s/l} \in \mathbb{R}^{H\times W\times C_{s/l}}$ with a size of $k \times k \times C_{s/l}$ can be acquired. For one sample $x_{hsi}^{p}$ in dimension-reduced HSI, it is first sent to a group of 3-D convolutional layers, which can be formulized as

$$Conv_{3d_n}(x) = \boldsymbol{w}_{3d_n}\boldsymbol{x} + b_{3d_n} \tag{1}$$

$$ReLU_L(x) = \begin{cases} x, & x \geq 0 \\ \dfrac{x}{\alpha} & x < 0 \end{cases} \tag{2}$$

$$BN(x_i) = \gamma \times \left( \frac{x_i - \mu_1(x_i)}{\sqrt{\sigma(x_i)^2 + \varepsilon}} \right) + \beta \tag{3}$$

$$f_n = BN(ReLU_L(Conv_{3d_n}(\boldsymbol{x}))) \quad (n = 1,2,3) \tag{4}$$

$$Conv_{2d_n}(x) = \boldsymbol{w}_{2d_n}\boldsymbol{x} + b_{2d_n} \tag{5}$$

$$\boldsymbol{o}_1 = ReLU_L(Conv_{2d_n}(t(f_3(f_2(f_1(x)))))) \tag{6}$$

where $\boldsymbol{x}$ denotes the input data, and $\boldsymbol{w}_{3d_n}$, $b_{3d_n}$ represent the weight matrix and the corresponding bias in the $n$th 3-D convolutional layer, respectively. $Conv_{3d_n}(\bullet)$ is a 3-D convolutional function ($n$ = 1, 2, 3 separately refers to the 3-D convolutional operation with a kernel size of 7 × 3 × 3, 5 × 3 × 3 and 3 × 3 × 3). $ReLU_L(\bullet)$ is LeakyReLU activation function. $\alpha$ indicates a parameter in $(1,+\infty)$. $BN(\bullet)$ denotes batch normalization, $x_i$ is the input batch, $\mu_1(\bullet)$ is an average calculation of the same layer among samples in one batch, $\sigma(\bullet)^2$ is variance calculation. $\gamma$ and $\beta$ are learnable scale parameter and shift parameter, respectively. $\varepsilon$ is a very small value introduced to prevent division by zero. $f_n(\bullet)$ is a group of 3-D convolution-associated operations. $t(\bullet)$ implies reshape operation that unites the second and third dimensions of the output of the last layer. $Conv_{2d_n}(\bullet)$ is a 2-D convolutional function ($n$ = 1 and 2 refer to the 2-D convolutional function with a kernel size of 3 × 3 and 1 × 1), and $\boldsymbol{w}_{3d_n}$, $b_{3d_n}$ represent the weight matrix and the corresponding bias in the $n$th 2-D convolutional layer. $o_1$ denotes the output of the above operations for HSI. With respect to the LiDAR image, if the number of channels $\mathrm{C_L}$ is greater than 1, the average calculation is first taken for the channel dimension. While the number of channels $\mathrm{C_L}$ is equal to 1, the original LiDAR data is directly the input. The output of the above operations $o_2$ can be expressed as.

$$p(x) = \begin{cases} \mu_2(x), & C_L > 1 \\ x & C_L = 1 \end{cases} \tag{7}$$

$$g(x) = BN(ReLU_L(Conv_{2d_1}(x))) \tag{8}$$

$$\boldsymbol{o}_2 = g(g(p(x))) \tag{9}$$

where $p(\bullet)$ is the preprocessing operation, $\mu_2(\bullet)$ is the mean calculation of the channel dimension. $g(\bullet)$ is a series of 2-D convolution-related calculations.

After getting $o_1$ and $o_2$, a multimodal heterogeneous graph encoder is followed to extract representatively multi-sensor features. Define $m(x)$ is the mask generation function, $\sigma(\bullet)$ is the sigmoid function, and $\bullet$ is matrix multiplication. $\boldsymbol{T}$ is gained to prepare the input of GCN by multiplying the mask and the post-convolution value, which is helpful to exclude redundant features. In heterogeneous scenarios, not all patch elements are equally informative. This step acts as a learnable saliency gate on graph nodes, and the gated features are used for subsequent relation modeling and dynamic adjacency construction. In this way, background or low-response nodes are down-weighted before building the heterogeneous graph, so that later graph operations focus on structurally reliable regions. Besides, dense graph on all patch elements is a direct alternative. However, in heterogeneous HSI–SAR/LiDAR scenes, some nodes correspond to background, mixed pixels, or sensor-specific artifacts. A GCN operating on a dense graph tends to aggregate information from these low-informative or noisy neighbors, which increases over-smoothing and makes the model more sensitive to noise. One natural way is to use a fixed structure that does not depend on the input features. Such a mask is not learnable and applies the same connectivity prior to all patches, regardless of their content or cross-modal discrepancies, so it is difficult to adaptively suppress noisy or background nodes. Another approach is to compute a hand-crafted mask from part features. It depends on rules and thresholds that are non-learnable during training. This reliance on dataset- and modality-specific constraints makes the approach less stable, preventing it from capturing complex class-specific or cross-modal dependencies [48].

$\boldsymbol{Q}$, $\boldsymbol{K}$, and $\boldsymbol{V}$ represent Query, Key, and Value after dimensional reconstruction to match subsequent calculations in the multimodal heterogeneous graph encoder. The former sets the channel dimension back, the latter puts the channel dimension forward. $s(\bullet)$ represents a softmax function, $\boldsymbol{x}_i$ is the $i$th elements of the input, and $N$ is the number of the elements in the input. $\boldsymbol{A}$ is the content-specific attention map. $Conv_{1d_n}(\bullet)$ denotes a 1-D convolutional function ($n = 1$ stands for the kernel size of 1 × 1) for dynamic weight reconstruction in GCN. $\boldsymbol{W}$ is the updated weight to quantify the contribution of features and $\boldsymbol{M}_r$ represents a relationship matrix for encoding content-aware adjacency between the input. In particular, only the graph representation of HSI is adopted for subsequent feature learning as HSI contains more spectral-spatial information. Given an input patch with a patch size $k$, we regard each spatial location as a graph node, resulting in $k^2$ nodes. Each spatial location within the patch is treated as a node and used to construct a content-adaptive graph for information transformation. This process can be summarized in four steps and the input-specific salient non-Euclidean graph representation $\boldsymbol{G}$ can be deduced as Eqs. (10)–(17):

1) Saliency masking: we generate a soft mask via a lightweight convolution followed by a sigmoid (Eqs. (10)–(11)) to suppress background or low-response nodes.

2) Feature gating and reliability modulation: the masked feature is obtained by element-wise gating (Eq. (12)), and a reconstructed response gate is further computed to emphasize reliable structures (Eq. (13)).

3) Content-specific relations: we compute an attention map by applying softmax to the similarity between projected features (Eqs. (14)–(15)), where content-specific attention map encodes pairwise node relations and acts as the graph connectivity.

4) Dynamic weighting and aggregation: a dynamic weight is generated from the intermediate propagation using a 1-D convolution (Eq. (16)), and the final graph representation is produced by jointly applying relation-based aggregation, dynamic weighting, and the reliability gate (Eq. (17)).

$$\sigma(x) = \frac{1}{1+e^{-x}} \tag{10}$$

$$m(x) = \sigma(Conv_{2d_2}(\boldsymbol{x})) \tag{11}$$

$$\boldsymbol{T} = Conv_{2d_2}(\boldsymbol{x}) \times m(\boldsymbol{x}) \tag{12}$$

$$\boldsymbol{M}_r = \sigma(\boldsymbol{K} \bullet \boldsymbol{T}) \tag{13}$$

$$s(x_i) = \frac{\exp(\boldsymbol{x}_i)}{\sum_{j=1}^{N} \exp(\boldsymbol{x}_j)} \tag{14}$$

$$\boldsymbol{A} = s(\boldsymbol{Q} \bullet \boldsymbol{K}) \tag{15}$$

$$\boldsymbol{W} = Conv_{1d_1}(\boldsymbol{A} \bullet \boldsymbol{V}),\ \boldsymbol{A} = s(\boldsymbol{Q} \bullet \boldsymbol{K}) \tag{16}$$

$$\boldsymbol{G} = (\boldsymbol{A} \bullet \boldsymbol{V}) \bullet \boldsymbol{W} \bullet \boldsymbol{M}_r \tag{17}$$

In this design, the adjacency for each patch is constrained by spatial proximity in terms of which nodes are allowed to be connected and is learned from multimodal features in terms of the edge strengths, yielding an input-specific heterogeneous graph. In particular, spectral-spatial relations are modeled in this adjacency matrix, whereas category-level relations are encoded in the class-discriminative node features learned on top of the same graph.

Typically, conventional GCNs rely on a fixed adjacency pattern (e.g., grid- or distance-based neighbors) or globally shared graph, which could indiscriminately aggregate irrelevant or noisy nodes and blur class boundaries in complex multimodal remote-sensing scenes. In contrast, the proposed approach is input-specific and content-adaptive. The soft mask suppresses redundant and background responses, while the attention map provides a feature-driven relation strength that serves as content-aware connectivity. Moreover, the dynamic weight further modulates the feature interaction, and the reconstructed response gate enhances reliable structures. As a result, the deduced representation enables more semantically consistent aggregation, improves robustness against noisy neighbors, and better captures irregular boundaries and cross-modal heterogeneity, yielding a more discriminative non-Euclidean representation for classification.

### *B. Multi-convolutional Modulator*

Long-range dependencies are essential in identifying the various land covers, especially for large-scale objects, such as grass, residential, and highway. The emergence of a transformer with MSA provides a solution for non-local semantic perception. However, modal divergence is quite hard to detect via undifferentiated global computation in MSA, but instead, brings a lot of non-essential features and calculations. So, a question worth pondering arises, how to model long-term dependencies with less redundant features and parameters? The proposed multi-convolutional modulator may offer new insight into mitigating the above limitations.

The design is conceived in accordance with the design criteria of the transformer. The multi-convolutional modulator is an encoder with 5 layers (depth=5). Before being passed into the multi-convolutional modulator, the patch-to-embedding operation is applied to produce input embedding vectors [46]. Concretely, linear projection is utilized via a projection matrix for all input patches to generate initial embeddings, meanwhile, an additional learnable class embedding is added and optimized during training. After that, class embedding and initial embeddings are concatenated to form final embeddings. To take location information into account, the learned position embeddings that can be automatically updated during the learning process of the network are attached for the input of a multi-convolutional modulator $\boldsymbol{o}_3$ . The mathematical expression of the above processing is as follows

$$\boldsymbol{o}_3 = \boldsymbol{E}_{class} \,||\, (\boldsymbol{E}_i \times \boldsymbol{M}_p) + \boldsymbol{E}_{pos} \tag{18}$$

where $\boldsymbol{E}_i$ derives input patches, $\boldsymbol{M}_p$ is projection matrix, $\boldsymbol{E}_{class}$ deduces learnable class embedding, || represents concatenation operation, and $\boldsymbol{E}_{pos}$ is the position embeddings. The detailed architecture of the proposed multi-convolutional modulator is shown in Fig. 2. With the input $\boldsymbol{o}_3$, the feature extraction process can be denoted as

$$\delta(\boldsymbol{x}) = 0.5\boldsymbol{x} + 0.5\boldsymbol{x} \bullet erf\left(\frac{\boldsymbol{x}}{\sqrt{2}}\right), \quad erf(x) = \frac{2}{\sqrt{\pi}}\int_0^x e^{-t^2}dt \tag{19}$$

$$\boldsymbol{o}_4 = Conv_{2d_1}(\delta(Conv_{2d_2}(\boldsymbol{o}_3))) \tag{20}$$

$$\boldsymbol{o}_5 = Conv_{2d_2}(\boldsymbol{o}_4 \odot Conv_{2d_2}(\boldsymbol{x})) \tag{21}$$

where $\delta(\bullet)$ is Gaussian error linear unit [49], $erf(\cdot)$ indicates error function [50], and $\odot$ stands for the Hadamard product. $\boldsymbol{o}_4$ and $\boldsymbol{o}_5$ individually refer to the result of the left branch and the final output. The arrangement of two branches facilitates simultaneous handling in the spatial and channel domains. Integrated design of the backbone behind these branches aims to focus on meaningful features while preserving information diversity. In the form of an elaborate pipeline of convolutional blocks, the proposed structure enables simple, tailored, and integrated encoding of high-level semantics.

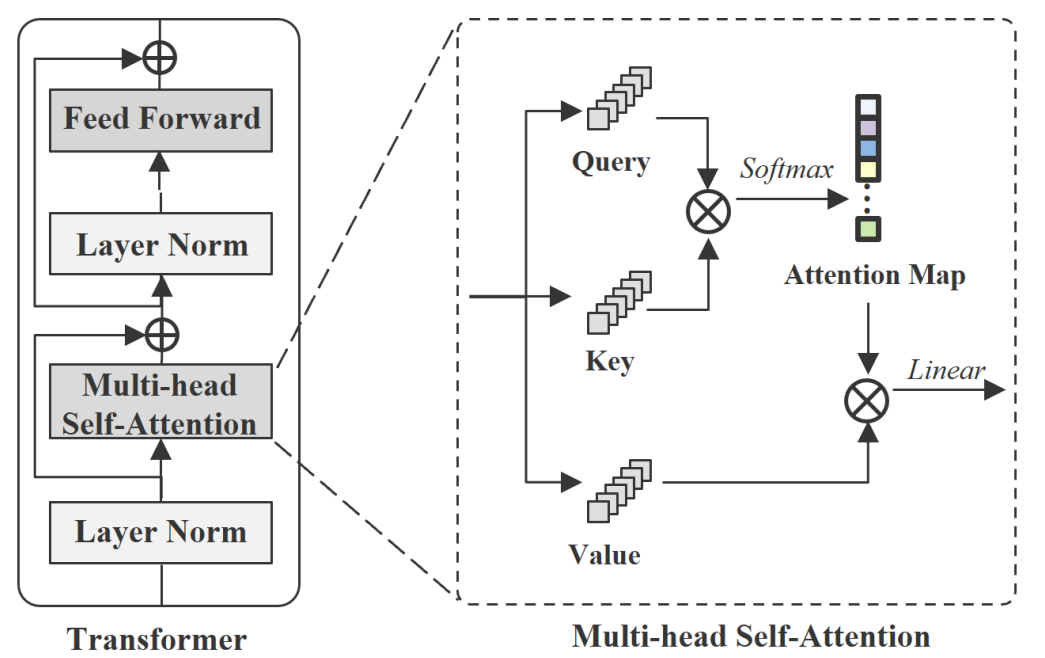

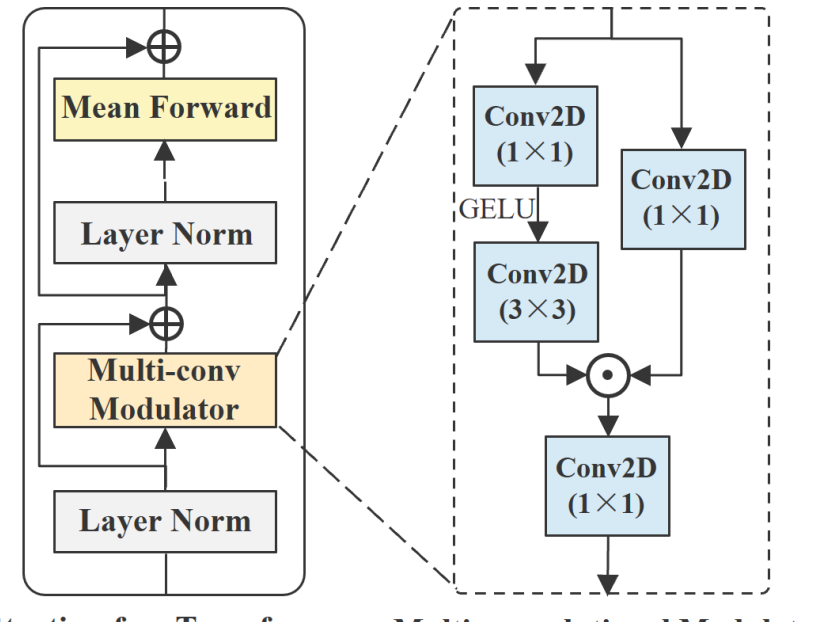

Fig. 3. Differences between the common transformer and the proposed method.

For a better comparison between the existing transformer and the proposed approach, Fig. 3 is employed to visually depict the differences. It is worth mentioning that an ordinary transformer performs calculations on all input features at each layer to capture long-range information, adding redundant features and additional calculations. In addition, the higher score of similar input patches in the attention map leads the network to focus more on analogous regions rather than geographically truly relevant features. Instead of the design paradigm of a conventional transformer, the proposed multi-convolutional modulator only modulates the long-term distinguishable information of the input patches from both spatial and channel dimensions via a simple convolutional computation. Due to the advantage of local connectivity in CNN, location properties are also more prominent in the proposed method. With multi-layer convolutions, long-distance geographical dependencies and inductive bias can be better modeled. Accordingly, the proposed structure can yield better performance than the traditional transformer [51].

Unlike the multi-convolutional modulator, which has a weight-sharing attribute within each channel, global calculations of MSA inevitably impose many parameters, resulting in efficiency reduction. The traditional MSA can be formulated below

$$Attn(q,k,v) = s(\frac{qk^T}{\sqrt{d}})v \tag{22}$$

where $\boldsymbol{q}$, $\boldsymbol{k}$, and $\boldsymbol{v}$ mean query, key, and value in the traditional MSA, respectively. $\boldsymbol{d}$ is the dimension of $\boldsymbol{q}$ and $\boldsymbol{k}$. The common transformer tends to flatten the input patch, then compute the global correlation score for the attention map of all inputs. Moreover, the universal attention map and value are performed by matrix multiplication to obtain the output of MSA. In the above process, all three parameters $\boldsymbol{q}$, $\boldsymbol{k}$, and $\boldsymbol{v}$ are converted from the same input and then involved in the calculation one by one, and are thus accompanied by overlapping features. As these three values can be regarded as the results attained through a series of linear projection, they can be expressed in a simplified way (i.e., omitting the bias) as

$$\boldsymbol{q},\ \boldsymbol{k},\ \boldsymbol{v} = Xw_q,\ Xw_k,\ Xw_v \tag{23}$$

where $w_q$, $w_k$, and $w_v$ represent the corresponding projection matrices of $\boldsymbol{q}$, $\boldsymbol{k}$, and $\boldsymbol{v}$. Then, Eq. (22) can be denoted as

$$Attn(\boldsymbol{q},\boldsymbol{k},\boldsymbol{v}) = s(\frac{Xw_q w_k^T X^T}{\sqrt{d}})v = s(\frac{XwX^T}{\sqrt{d}})(Xw_v) \tag{24}$$

where $w$ indicates the result of $(w_q w_k^T)$. In a similar way, Eq. (21) after omitting the bias for simplification could be overwritten as:

$$o_5 = W_4((W_2\delta(W_1X))(W_3X)) = W'\delta(W_1X)(W_3X) \tag{25}$$

where $W_i$ is the weight of the convolutional kernel used in a multi-convolutional modulator ($i$ = 1, 2, 3, 4 individually represents the top, bottom layer in the left branch, the layer in the right branch, and the layer in the ground layer in the main stem shown in Fig. 3), and ($W_4W_2$) can be replaced by $W'$. Interestingly, these two forms are equivalent for achieving remote semantic modeling. However, Eq. (24) has one more input $X$ to the calculation before the nonlinear activation function than Eq. (25), and one less $W'$ than Eq. (25). Specifically, $W'$ is the result of multiplying the weights of the 3 × 3 convolution kernel and the 1 × 1 convolution kernel. While $X$ stands for the input of the entire patch, whose size is much larger than $W'$, naturally brings more computational overhead. In terms of complexity, assume that the dimension of input is $N \times D$, the computation cost of the MSA can be derived as $O_{MSA}(N^2D + ND^2)$, while that of a CNN is $O_{Proposed}(NWD^2)$ ($\sqrt{W}$ is the convolutional kernel size). Apparently, MSA requires more computational cost. Nonetheless, the performance gain from MSA is still capped as a result of numerous feature similarity calculations.

Different from classical multi-branch modules, the proposed multi-convolutional modulator is explicitly designed to operate on graph-encoded, multimodal features and to act as a feature recalibration stage. First, the input features already embed heterogeneous graph structure, therefore the modulator focuses on re-weighting spectral-spatial patterns that are informed by the learned adjacency instead of re-extracting generic texture features. Second, the number of branches and channel widths are constrained so that the additional parameters are modest compared with stacking deeper GCN or CNN layers.

In addition, standard multi-head self-attention (MSA) enables global token interactions and is effective for modeling long-range dependencies. However, in multimodal HSI–SAR/LiDAR classification, directly relying on global attention for token mixing can be suboptimal. First, MSA performs global, content-agnostic mixing in the sense that every token is allowed to interact with all others, which may propagate redundant or modality-inconsistent cues when the two modalities exhibit heterogeneous noise patterns and distinct spatial responses. Second, under limited supervision, the flexibility of global attention may amplify spurious correlations and reduce robustness across scenes, especially when long-range interactions dominate local boundary cues. Compared with standard MSA, the multi-convolutional modulator offers three points: (1) structured long-range modeling with locality preservation: stacked convolutions progressively enlarge the effective receptive field, enabling long-range context integration while preserving local continuity and boundaries; (2) noise suppression and stable multimodal refinement: convolutional mixing provides an inductive bias that acts as a structured filter, mitigating excessive global feature diffusion and reducing modality-specific artifacts before final classification; (3) complementary role to the variably salient graph: after the heterogeneous graph encoder constructs a variably salient graph representation, the modulator refines the resulting token embeddings using locality-aware operations, improving discriminability and stability for multimodal feature learning.

Overall, MSA exploits the input data multiple times, bringing in both a lot of redundant features and computational costs. The proposed multi-convolutional modulator can encode long-range property distribution with less cost. Thus, the proposed approach is more promising than the common transformer regarding performance and efficiency.

### C. Mean Forward

HSI labeling is extremely difficult due to variations in environment, lighting, and imaging conditions. That means there is a scarcity of labeled data that can be served as training samples, so the overfitting phenomenon is particularly evident. To mitigate overfitting, we expect the network not to be sensitive to only one or a few categories of features, but to learn patterns among a multitude of ones. The introduction of the average can achieve this goal by counteracting the effects of particular points or outlier classes and enhancing the model's overall characterization of the data. In addition, different networks may produce varying overfitting, and the average enables some "opposite" fits to cancel each other out. Based on the above considerations, a mean forward is presented to optimize network training and mitigate overfitting.

Assume $\boldsymbol{o}_{6,i}$ denotes the $i$th vector of output $\boldsymbol{o}_6$, the mean forward can be deduced as

$$\boldsymbol{o}_6 = f(\delta(f(\boldsymbol{x}))) \tag{26}$$

$$\boldsymbol{o}_{6,i} = 0.5(avg(\boldsymbol{o}_6) + \boldsymbol{o}_{6,i}) \tag{27}$$

where $\boldsymbol{x}$ is input, $f(\bullet)$ represents linear projection, and $avg(\bullet)$ stands for the average function. In this simple way, mean features that can represent non-local data properties are supplemented into the network stream, making the network focus less on anomalous attributes or noisy features and tend to fit the overall data instead. Further, if one part of the network is over-fitted for HSI and another part is excessively learned for SAR or LiDAR image, the addition of mean feature representation facilitates the alleviation of internal biases, allowing cross-modal learning to concentrate more on exploiting inter-modal consistency and complementarity. Also, the supervision of the average signals can relieve the effect of redundant features.

Therefore, a simple yet competitive mean forward is designed for reinforcing holistic data representation. With the compensation of average values, the network can map output features to a more comprehensive representation for HSI-SAR/LiDAR joint classification, easing the overfitting phenomenon of the training process with sparse annotations [52].

Compared with explicit feature fusion, the proposed mean forward approach incurs only a marginal increase in computation and no additional learned parameters. It can be viewed as an implicit ensemble tailored to the proposed framework: the multimodal heterogeneous graph encoder and multi-convolutional modulator generate rich and structurally informed features, while mean forward aggregates multiple slightly perturbed views of these features to produce more reliable results.

On the basis of these three structures, a THSGR approach is established for heterogeneous classification. The presented model is trained via an end-to-end strategy with the adaptive moment estimation (Adam) optimizer [53], and the cross-entropy loss function $L_{CE}$ can be calculated as

$$y_{ic} = \begin{cases} 1, & c = Y_i \\ 0, & c \neq Y_i \end{cases} \tag{28}$$

$$L_{CE} = -\frac{1}{N}\sum_{i=1}^{N}\sum_{c=1}^{C}(y_{ic}\log \hat{y}_{ic}) \tag{29}$$

where $y_{ic}$ represents the targeted output of the $i$th training sample for the class $c$, $\hat{y}_{ic}$ stands for the predicted probability result generated by the softmax activation function. $Y_i$ denotes the truth label of the $i$th training sample. $N$ is the batch size, and $C$ is the number of categories. Through the loss function $L_{CE}$ and backpropagation [54], the model can be optimized for more complete, more exact, and finer classification performance.

## III. Experiments and Analysis

In this section, three benchmark HSI-SAR/LiDAR datasets are employed. Subsequently, the experimental results and the associated analysis are given.

### *A. Dataset Description*

Three widely used HSI scenarios and their corresponding SAR or LiDAR data with different modalities, resolutions, and sampling strategies are utilized in the experiments. The first dataset is HSI-SAR Augsburg, which is composed of 332 × 485 pixels. The HSI includes 180 spectral channels, while the SAR data has four features. The second one is HSI-LiDAR Houston 2013, which consists of 349 × 1905 pixels. The HSI covers 144 spectral bands, and the LiDAR contains one band. The third dataset is HSI-SAR Berlin with 1723 × 476 pixels. The number of bands is 244 and 4 for HSI and SAR, respectively. The first and third datasets can be downloaded from https://github.com/danfenghong/ISPRS_S2FL, and the second dataset is available at https://hyperspectral.ee.uh.edu/?page_id=459 via data request. The pseudo-color images of HSI, SAR/LiDAR, and ground-truth in three datasets are illustrated in Figs. 5–7.

To comprehensively evaluate the classification performance, random sampling and spatially isolated sampling are employed in this study. In Augsburg and Houston 2013 datasets, 50 samples of each category are randomly chosen from ground-truth to form the training set, and the remaining samples are used as the test set. In the Berlin dataset, spatially disjoint training-test sets are adopted as displayed in Fig. 4. The concrete number of training and test samples is tabulated in Table I.

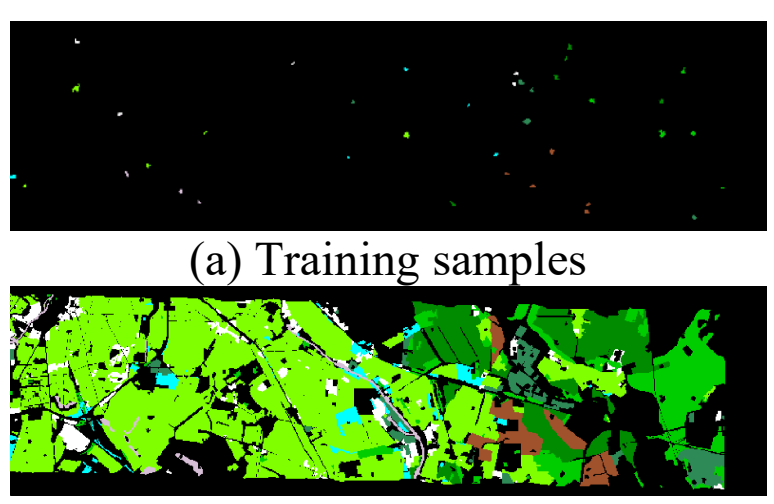

(a) Training samples

(b) Test samples

Fig. 4. Spatially disjoint training-test sets of Berlin dataset.

TABLE I
Training-Test Distribution of Berlin Dataset

| Class number | Class name | Training | Test |
|---|---|---|---|
| 1 | Forest | 443 | 54511 |
| 2 | Residential Area | 423 | 268219 |
| 3 | Industrial Area | 499 | 19067 |
| 4 | Low Plants | 376 | 58906 |
| 5 | Soil | 331 | 17095 |
| 6 | Allotment | 280 | 13025 |
| 7 | Commercial Area | 298 | 24526 |
| 8 | Water | 170 | 6502 |
| | Total | 2820 | 461851 |

### *B. Experimental Setup*

The proposed approach is performed on the PyTorch platform with an NVIDIA GeForce RTX 3090Ti GPU. Adam optimizer is utilized during training with an initial learning rate of 0.001 and weight decay of 0.001. A multi-step learning rate schedule is employed with a factor of 0.01 at 0.5 and 5/6 of the total epochs. The loss function is label smoothing cross entropy with a smoothing factor of 0.01. The epoch is empirically set to 200, 50, and 60 for Augsburg, Houston 2013 and the disjoint Berlin dataset, respectively.

In order to verify the effectiveness of the proposed approach, SOTA comparison algorithms are utilized for experiments, including Global-local transformer network (GLTNet) [55], Deep encoder-decoder network (EndNet) [56], Multimodal deep learning (MDL) framework [57], Shared and specific feature learning (S2FL) [58], L1-norm sparse regression-based common subspace learning (CoSpace-l1) [59], L2-regularized ridge regression-based common subspace learning (CoSpace-l2) [60], Multihead self-attention and graph convolution (MSA-GCN) [61], Graph-attention based multimodal fusion network (GAMF) [62], the presented single-modal HSI model (HSI), the designed single-modal SAR/LiDAR method (SAR/LiDAR), and THSGR. The performance of each network is evaluated in overall accuracy (OA), Kappa coefficient (Kappa), floating point operations (FLOPs), multiply accumulate operations (MACs), and parameters (Params).

### *C. Performance Comparison*

In order to thoroughly validate the classification capability, THSGR is compared not only with outstanding multimodal classification works but also with the single-modal form of the proposed method from both quantitative and qualitative perspectives in this section.

Tables II-IV list quantitative reports in three real-world HSI-SAR/LiDAR datasets (better results are bolded). Compared with multi-source data classification studies, the OA and Kappa of S2FL, CoSpace-l1, and CoSpace-l2 are unsatisfactory due to the shortage of deep features. To be more specific, S2FL learns linear regression matrix, shared and specific subspace projection matrices to capture cross-modal information. CoSpace-l1 builds an l1-penalty common subspace for modality-specific information, while CoSpace-l2 linearly learns a latent shared subspace via aligning the manifold structure of two modalities. The above-mentioned three approaches employ conventional subspace learning, making the acquired features more shallow, less representative, and less abstract. As a result, the performance is limited to a large extent. For instance, the accuracy of these models is 24%-36% inferior to that of the proposed method in Augsburg dataset with fewer than 0.5% of the training samples. EndNet, as the typical method based on CNN, yields worse, better, as well as closer OA and Kappa than three subspace-based methods in Augsburg, Houston 2013, and Berlin datasets, respectively. The reason for this phenomenon exists in the overfitting phenomenon. Since EndNet comprises multiple sophisticated encoder-decoder structures, the model is prone to overfitting the training data, leading to a sharp drop in the accuracy of test sets. Also, the small size of the dataset exacerbates this phenomenon, rendering the accuracy to be the lowest in Augsburg dataset, i.e., the OA of 51.45% and Kappa of 38.29%. Additionally, MDL is also designed through CNN, gaining higher accuracy than S2FL, CoSpace-l1, and CoSpace-l2 in Houston 2013 and Berlin datasets. In Augsburg dataset, the performance of MDL is better than that of CoSpace-l1, whereas worse than that of S2FL and CoSpace-l2. This change indicates that MDL is much less effective for small-scale datasets with alike objects due to the lack of salient features. Further, the inconsistent results of EndNet and MDL in various datasets also reflect that these approaches are less robust. GLTNet utilizes a transformer for modeling long-term dependencies and acquires better impact than EndNet, MDL, S2FL, CoSpace-l1, and CoSpace-l2 in all datasets. The factor can be attributed to the long-distance semantic extraction capability of the transformer. Nevertheless, long-range features may introduce redundant features that interfere with classification, making critical features inadequate and causing many misclassifications. As a consequence, the accuracy of GLTNet is much lower than the proposed algorithm in three datasets. Even more, the small training set also restricts the effectiveness of MSA. For GCN-based popular methods, MSA-GCN and GAMF, THSGR achieves an OA of 87.39%, which is 10.85% and 3.51% higher than MSA-GCN and GAMF on the Augsburg dataset, respectively. On the Houston 2013 dataset, THSGR reaches 97.09% OA and 96.85% Kappa, improving upon MSA-GCN by 1.62% and 1.75%, and upon GAMF by 6.03% and 6.52%. For the Berlin dataset, THSGR attains the highest OA of 67.75%, which is 0.73% and 6.65% higher than MSA-GCN and GAMF, while its Kappa is slightly lower than that of MSA-GCN but still higher than GAMF by 4.70%. These results indicate that, although MSA-GCN achieves a marginally better Kappa on Berlin, the proposed THSGR provides higher and more robust overall performance across all three datasets and yields much larger gains on Augsburg and Houston 2013 scenarios.

When it comes to considering the proposed multimodal THSGR with unimodal HSI, SAR, or LiDAR, it can be found that HSI with only HSI involved behaves much better than SAR/LiDAR with only SAR/LiDAR data engaged. This observation hints that HSI with numerous bands provides more information than SAR/LiDAR with very few channels. By contrast, the proposed THSGR is even better than HSI, confirming that the structural attributes and elevation information derived from SAR and LiDAR can assist in the classification. Thereby, the proposed heterogeneous approach yields higher accuracy than HSI or SAR/LiDAR. Especially in the spatially disjoint large-scale Berlin dataset, THSGR can still increase the OA by exceeding 6% and 18% compared to HSI and LiDAR, respectively.

From a holistic perspective, there is an abnormal phenomenon when evaluating all methods together: the proposed HSI is frequently more accurate than some heterogeneous algorithms. For example, HSI outperforms GLTNet by about 2% in Houston 2013 dataset. Even the low-performing SAR shows over 2% and 4% improvement in terms of OA over MDL and CoSpace-l1, respectively, in Augsburg dataset. This is primarily due to the lack of discriminative information in these methods, which leads to unreasonable feature fusion. Contrastingly, the proposed model can individually capture saliently category-specific signatures and geography-related contents via the multimodal heterogeneous graph encoder and multi-convolutional modulator. Therefore, the proposed THSGR is able to obtain better performance than other advanced methods in all datasets. Even in the hard-to-classify Houston 2013 dataset with just 50 training samples per class, THSGR gains the OA of 97.09%, which also embodies the power to mitigate the overfitting problem.

The qualitative classification maps and enlarged area are further supplied for visualization in Figs. 5–7. In contrast to the multi-sensor approach, it can be observed that Figs. 5–7(g)–(i) have many misclassifications whether for large-scale class 2 (residential area) in Augsburg and Berlin datasets or small-scale class 9 (road) in Houston 2013 dataset. This is because subspace-relevant algorithms struggle to exploit inter-modal consistency and complementarity, making it difficult to retrieve differentiated features that can contribute to classification. Due to the local information extraction ability of CNN, EndNet and MDL are susceptible to interference from small changes in the neighborhood. Consequently, the visualizations of EndNet and MDL suffer from a large amount of salt and pepper noise. This trend is particularly evident in Fig. 5(e)–(f) of Augsburg dataset, i.e., class 1 (forest) and class 2 (residential area) are often incorrectly classified into other similar categories such as class 3 (industrial area) and class 5 (allotment), resulting in a loss of geographical continuity of land covers. Conversely, there are fewer noisy points in the classification maps of GLTNet thanks to the considerations for non-local information of transformer, yet bringing detail missing and over-smoothing issue. Take the Fig. 6 of the spatially isolated Berlin dataset as an illustration, it can be seen that piecemeal distributed class 7 (commercial area) in class 2 (residential area) is not well distinguished, leaving the classification visualization of class 2 (residential area) too flat and not in line with reality. Regarding MSA-GCN and GAMF in Figs. 5–7(j)–(k), there is more noise and misclassifications, such as, class 5 (allotment), class 2 (residential area), and class 2 (residential area). Accordingly, the classification charts of the proposed approach are more consistent with ground-truth, covering not only geometrically structural characteristics but also clearly local details with less salt and pepper noise.

Regarding the proposed single-source HSI, SAR/LiDAR, and multi-source THSGR, it is worth mentioning that HSI gains better visualizations than SAR/LiDAR owing to rich spatial-spectral information. Notably, in the challenging Houston 2013 dataset, the classification maps of HSI involve fewer incorrect classifications, but for SAR/LiDAR the opposite. For example, the classification results of SAR/LiDAR inevitably carry a large number of mistakes even for categories with large inter-class variance, e.g., class 8 (commercial) and class 10 (highway) in Fig. 5. Nevertheless, with regard to land covers with comparable appearance such as class 1 (healthy grass) and class 4 (trees), structure or height information from SAR/LiDAR is desirable to assist in clarifying them. In comparison, the presented heterogeneous approach can conjoin spectral-spatial features from HSI and height or structural properties from SAR or LiDAR data to attain clearer, more accurate, and finer classification graphs.

As a whole, the proposed THSGR can attain better visualization whether in competition with advanced multimodal models or with the proposed single mode. The visualization results in Fig. 7 of the tough Berlin dataset with disjoint training and test set can serve as a representative example to support the aforementioned finding: the visualizations of the proposed approach are more exact even for difficult class 2 (residential area) and class 7 (commercial area) as a result of geographical dependencies and topological details and are more robust to the noise effect than all other methods.

The quantitative and qualitative comparisons follow the same trend, which can prove that with the careful layout of the proposed approach including multimodal heterogeneous graph encoder, multi-convolutional modulator and mean forward, native commonalities and vital differences between modalities, valuably semantic structures, as well as less overfitting can be tackled for refined land cover mapping.

TABLE II
QUANTITATIVE CLASSIFICATION RESULTS OF THE AUGSBURG DATASET.

| Class | GLTNet | EndNet | MDL | S2FL | CoSpace-l1 | CoSpace-l2 | MSA-GCN | GAMF | HSI | SAR | Proposed |
|---|---|---|---|---|---|---|---|---|---|---|---|
| 1 | **97.26 ±2.30** | 51.45 ±4.42 | 89.53 ±1.91 | 87.95 ±1.94 | 76.02 ±9.54 | 83.14 ±2.31 | 94.51 ±2.61 | 91.82 ±6.62 | 94.11 ±2.60 | 68.59 ±34.77 | 94.33 ±1.67 |
| 2 | 80.78 ±5.81 | 56.38 ±2.22 | 65.16 ±4.53 | 39.96 ±4.16 | 26.82 ±9.96 | 43.62 ±2.92 | 66.90 ±9.26 | 83.52 ±6.63 | 82.20 ±5.39 | 60.63 ±5.18 | **89.61 ±4.33** |
| 3 | 55.30 ±11.08 | 38.29 ±5.46 | 41.47 ±2.73 | 40.59 ±5.31 | 36.22 ±7.65 | 40.93 ±2.92 | 47.95 ±7.75 | 61.48 ±19.17 | **71.49 ±7.05** | 64.80 ±6.08 | 68.59 ±7.89 |
| 4 | 80.81 ±5.56 | 51.45 ±4.42 | 21.57 ±10.01 | 80.50 ±2.62 | 69.98 ±13.62 | 74.05 ±1.98 | 84.15 ±5.30 | 85.17 ±7.72 | **89.24 ±3.87** | 46.39 ±15.78 | 85.72 ±3.13 |
| 5 | 88.95 ±9.62 | 56.38 ±2.22 | 69.45 ±13.38 | 82.97 ±3.84 | 56.69 ±6.27 | 78.90 ±3.16 | 82.64 ±4.98 | 85.30 ±14.68 | **95.05 ±2.11** | 37.07 ±6.77 | 92.72 ±3.51 |
| 6 | 43.30 ±14.51 | 38.29 ±5.46 | 41.03 ±4.74 | 43.24 ±1.50 | 38.34 ±6.66 | 41.38 ±1.76 | 49.00 ±8.87 | 67.68 ±22.74 | **71.56 ±7.54** | 32.50 ±9.52 | 68.65 ±6.44 |
| 7 | 61.30 ±6.54 | 51.45 ±4.42 | 65.85 ±3.31 | 62.88 ±3.75 | 50.72 ±2.44 | 61.20 ±2.37 | 42.58 ±16.71 | 70.30 ±6.20 | **80.18 ±3.37** | 14.62 ±16.37 | 75.77 ±1.94 |
| OA (%) | 81.31 ±3.68 | 51.45 ±4.42 | 52.77 ±3.84 | 63.01 ±2.10 | 51.51 ±8.43 | 61.30 ±1.20 | 76.54 ±3.68 | 83.88 ±3.93 | 85.98 ±2.27 | 55.70 ±9.15 | **87.39 ±2.36** |
| Kappa (%) | 74.65 ±4.56 | 38.29 ±5.46 | 39.97 ±4.73 | 53.13 ±2.34 | 39.87 ±10.21 | 51.10 ±1.28 | 68.41 ±4.71 | 77.99 ±5.02 | 80.86 ±2.86 | 44.63 ±9.36 | **82.44 ±3.07** |

TABLE III
QUANTITATIVE CLASSIFICATION RESULTS OF THE HOUSTON 2013 DATASET.

| Class | GLTNet | EndNet | MDL | S2FL | CoSpace-l1 | CoSpace-l2 | MSA-GCN | GAMF | HSI | LiDAR | Proposed |
|---|---|---|---|---|---|---|---|---|---|---|---|
| 1 | 96.60 ±1.42 | 96.75 ±1.06 | **97.35 ±0.53** | 95.77 ±1.84 | 94.82 ±1.58 | 94.55 ±1.89 | 97.10 ±0.88 | 92.22 ±7.10 | 93.81 ±3.54 | 25.95 ±2.05 | 96.37 ±0.98 |
| 2 | 98.27 ±1.94 | 97.46 ±0.95 | 97.21 ±1.01 | 98.11 ±0.77 | 96.50 ±2.15 | 97.16 ±1.79 | 98.92 ±1.07 | 94.10 ±6.30 | 98.80 ±1.45 | 14.39 ±4.56 | **99.20 ±0.59** |
| 3 | 99.20 ±1.11 | 99.94 ±0.08 | 99.81 ±0.34 | **100.00 ±0.00** | **100.00 ±0.00** | 99.72 ±0.34 | 99.85 ±0.12 | 98.68 ±3.63 | 99.88 ±0.06 | 75.80 ±19.96 | 99.88 ±0.06 |
| 4 | **98.34 ±1.57** | 94.71 ±1.42 | 95.53 ±0.69 | 97.64 ±0.35 | 95.83 ±0.59 | 94.00 ±1.53 | 97.94 ±2.31 | 95.76 ±2.89 | 97.29 ±3.31 | 75.04 ±6.45 | 96.47 ±2.37 |
| 5 | 99.78 ±0.35 | 98.84 ±0.49 | 98.98 ±0.23 | 99.48 ±0.17 | 98.64 ±0.53 | 97.15 ±1.53 | **99.94 ±0.13** | 97.68 ±2.20 | 99.19 ±1.00 | 41.59 ±12.62 | 99.31 ±0.80 |
| 6 | 98.18 ±2.26 | 96.87 ±0.61 | 97.09 ±1.15 | 99.49 ±0.20 | 97.96 ±1.62 | 97.02 ±1.86 | 98.54 ±0.92 | 98.50 ±1.62 | 99.13 ±1.07 | 60.07 ±12.61 | **100.00 ±0.00** |
| 7 | 88.78 ±3.60 | 90.11 ±1.30 | 89.75 ±2.09 | 91.38 ±4.36 | 90.53 ±3.16 | 88.62 ±3.94 | 92.48 ±3.00 | 94.67 ±3.44 | **96.24 ±1.59** | 47.64 ±8.15 | 93.68 ±2.13 |
| 8 | 94.12 ±3.98 | 95.29 ±0.51 | 93.97 ±1.30 | 85.16 ±2.32 | 79.08 ±3.38 | 90.87 ±2.96 | **96.06 ±0.94** | 86.18 ±4.90 | 87.92 ±2.57 | 62.46 ±11.85 | 93.57 ±3.63 |
| 9 | 85.47 ±10.32 | 80.13 ±3.84 | 79.78 ±1.42 | 75.52 ±4.49 | 63.84 ±5.55 | 77.17 ±2.66 | 83.74 ±8.76 | 85.81 ±10.33 | 91.11 ±2.86 | 19.12 ±3.53 | **92.33 ±3.18** |
| 10 | 91.79 ±8.35 | 91.11 ±3.28 | 92.54 ±2.17 | 86.68 ±5.10 | 67.75 ±1.68 | 91.13 ±2.05 | 96.44 ±4.53 | 77.32 ±22.77 | **99.93 ±0.14** | 25.69 ±12.27 | 99.34 ±0.79 |
| 11 | 95.48 ±0.99 | 95.86 ±1.60 | 95.65 ±2.07 | 86.33 ±2.11 | 82.30 ±3.82 | 90.36 ±1.80 | 97.14 ±4.57 | 87.10 ±15.52 | **99.27 ±0.87** | 82.16 ±6.78 | 98.94 ±0.52 |
| 12 | 86.05 ±10.93 | 91.45 ±3.60 | 88.76 ±2.34 | 86.68 ±2.76 | 76.45 ±1.75 | 82.33 ±4.12 | 90.06 ±4.71 | 85.34 ±20.29 | 93.42 ±2.82 | 16.45 ±13.79 | **97.01 ±3.42** |
| 13 | 94.94 ±1.53 | 71.50 ±1.27 | 72.89 ±2.78 | 68.35 ±2.69 | 57.14 ±5.09 | 60.38 ±3.95 | 90.96 ±4.58 | 94.82 ±1.06 | **97.80 ±3.02** | 65.35 ±6.83 | 97.76 ±2.12 |
| 14 | 99.89 ±0.14 | 98.78 ±0.24 | 98.25 ±0.24 | 99.42 ±0.34 | 98.68 ±1.48 | 99.21 ±0.00 | 99.97 ±0.08 | **100.00 ±0.00** | **100.00 ±0.00** | 78.73 ±5.93 | **100.00 ±0.00** |
| 15 | 99.77 ±0.32 | 99.97 ±0.07 | **100.00 ±0.00** | 99.05 ±0.44 | 99.11 ±0.25 | 99.11 ±0.68 | **100.00 ±0.00** | 99.08 ±2.07 | 99.97 ±0.07 | 63.61 ±23.86 | **100.00 ±0.00** |
| OA (%) | 94.30 ±1.14 | 93.35 ±0.32 | 93.19 ±0.19 | 90.88 ±0.47 | 85.76 ±0.82 | 90.62 ±0.88 | 95.47 ±1.89 | 91.06 ±5.09 | 96.31 ±0.69 | 45.65 ±2.78 | **97.09 ±0.49** |
| Kappa (%) | 93.83 ±1.23 | 92.80 ±0.35 | 92.63 ±0.20 | 90.13 ±0.51 | 84.58 ±0.90 | 89.85 ±0.95 | 95.10 ±2.05 | 90.33 ±5.51 | 96.00 ±0.74 | 42.00 ±2.91 | **96.85 ±0.53** |

TABLE IV
QUANTITATIVE CLASSIFICATION RESULTS OF THE BERLIN DATASET.

| Class | GLTNet | EndNet | MDL | S2FL | CoSpace-l1 | CoSpace-l2 | MSA-GCN | GAMF | HSI | SAR | Proposed |
|---|---|---|---|---|---|---|---|---|---|---|---|
| 1 | 48.98 ±15.65 | 76.51 ±2.50 | 76.29 ±5.28 | 80.15 ±0.95 | **85.03 ±0.51** | 79.75 ±1.81 | 77.97 ±4.57 | 71.79 ±10.82 | 64.98 ±9.98 | 45.19 ±37.43 | 64.73 ±3.93 |
| 2 | 66.78 ±7.88 | 62.34 ±6.11 | 67.85 ±6.62 | 51.72 ±1.89 | 58.12 ±2.32 | 48.64 ±4.82 | 70.03 ±7.59 | 56.32 ±13.09 | 66.41 ±19.17 | 66.05 ±23.82 | **74.89 ±11.06** |
| 3 | 50.06 ±8.35 | 53.44 ±3.34 | 48.32 ±4.00 | 47.05 ±1.68 | 49.03 ±1.71 | 51.09 ±2.70 | 55.42 ±6.73 | 56.87 ±9.53 | **60.06 ±18.07** | 43.53 ±17.91 | 57.24 ±10.60 |
| 4 | **85.79 ±5.93** | 47.98 ±10.62 | 57.18 ±2.22 | 66.28 ±0.79 | 75.39 ±0.88 | 68.49 ±1.81 | 63.06 ±10.18 | 78.44 ±5.47 | 55.09 ±9.95 | 16.34 ±19.98 | 63.00 ±9.54 |
| 5 | **85.60 ±8.17** | 72.18 ±4.88 | 69.96 ±3.57 | 82.74 ±0.17 | 80.11 ±0.10 | 71.03 ±0.43 | 74.92 ±8.27 | 83.42 ±2.99 | 55.92 ±8.95 | 27.74 ±35.71 | 64.92 ±6.75 |
| 6 | 70.72 ±10.65 | 59.12 ±6.20 | 51.49 ±12.16 | 58.62 ±1.02 | 54.02 ±1.53 | 68.06 ±2.08 | **79.37 ±4.63** | 56.81 ±18.44 | 56.71 ±6.24 | 4.49 ±4.73 | 61.39 ±5.38 |
| 7 | 29.94 ±4.25 | 31.85 ±4.88 | 30.91 ±2.97 | 25.66 ±1.02 | 27.05 ±1.30 | 31.35 ±0.98 | 35.44 ±7.68 | **37.66 ±10.73** | 24.13 ±10.31 | 1.70 ±2.00 | 26.99 ±14.97 |
| 8 | 61.59 ±6.97 | 59.62 ±3.31 | 58.67 ±2.82 | 55.01 ±0.83 | 58.58 ±0.64 | 59.59 ±1.59 | **64.83 ±9.34** | 62.20 ±15.99 | 44.88 ±6.17 | 14.45 ±20.74 | 46.63 ±8.21 |
| OA (%) | 65.19 ±3.47 | 60.43 ±3.29 | 64.20 ±2.89 | 56.75 ±1.08 | 62.18 ±1.29 | 55.56 ±2.63 | 67.02 ±3.81 | 61.10 ±7.62 | 61.32 ±9.91 | 49.02 ±9.40 | **67.75 ±5.58** |
| Kappa (%) | 51.49 ±3.07 | 45.02 ±3.07 | 48.37 ±1.77 | 42.64 ±0.90 | 48.33 ±1.15 | 42.09 ±2.09 | **54.69 ±4.87** | 48.35 ±7.27 | 45.76 ±7.38 | 25.59 ±8.98 | 53.05 ±4.98 |

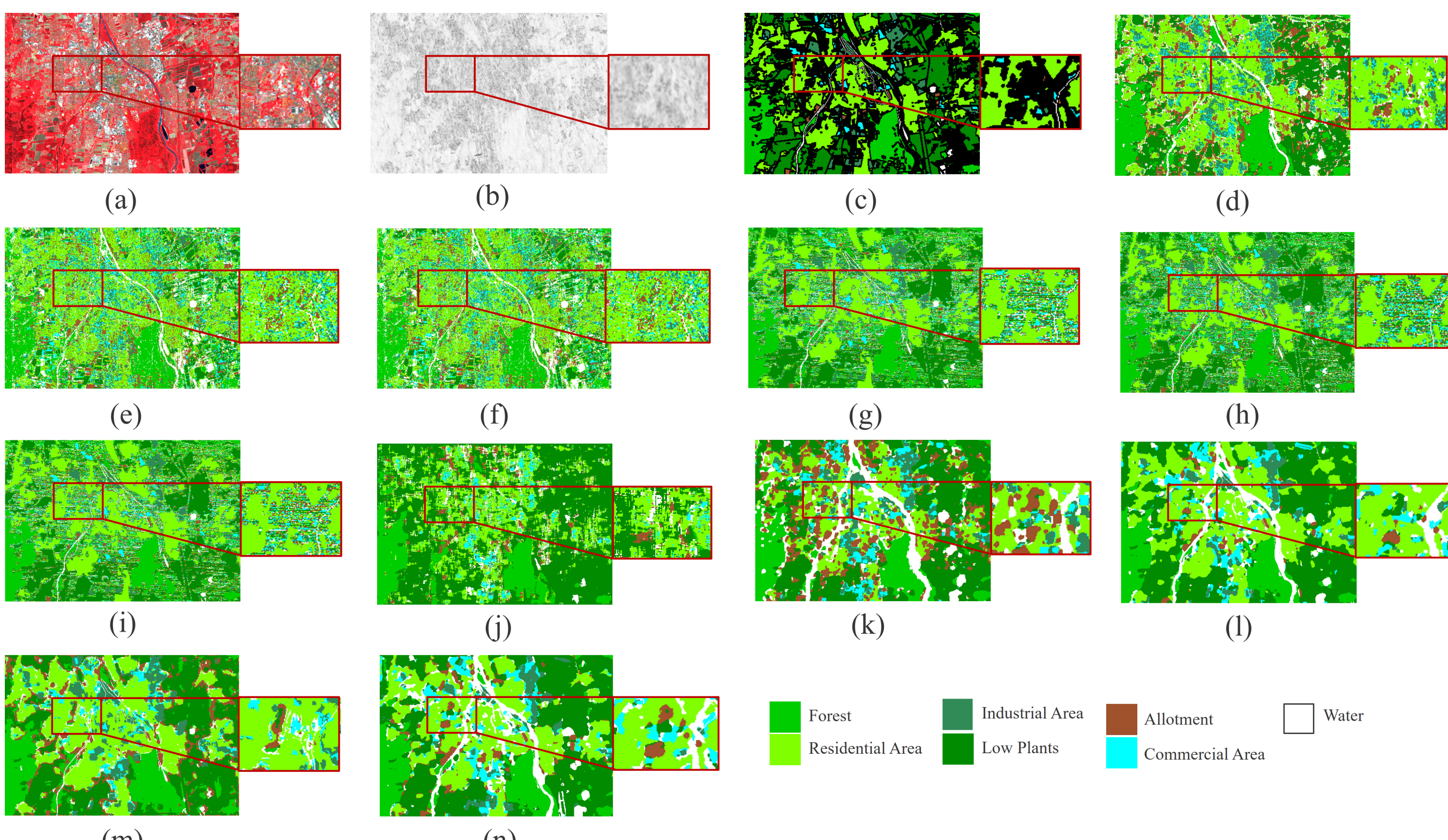

Fig. 5. Classification maps for Augsburg dataset. (a) Pseudo-color HSI. (b) SAR imagery. (c) Ground-truth map. (d) GLTNet. (e) EndNet. (f) MDL. (g) S2FL. (h) CoSpace-l1. (i) CoSpace-l2. (j) MSA-GCN. (k) GAMF. (l) HSI. (m) SAR. (n) Proposed.

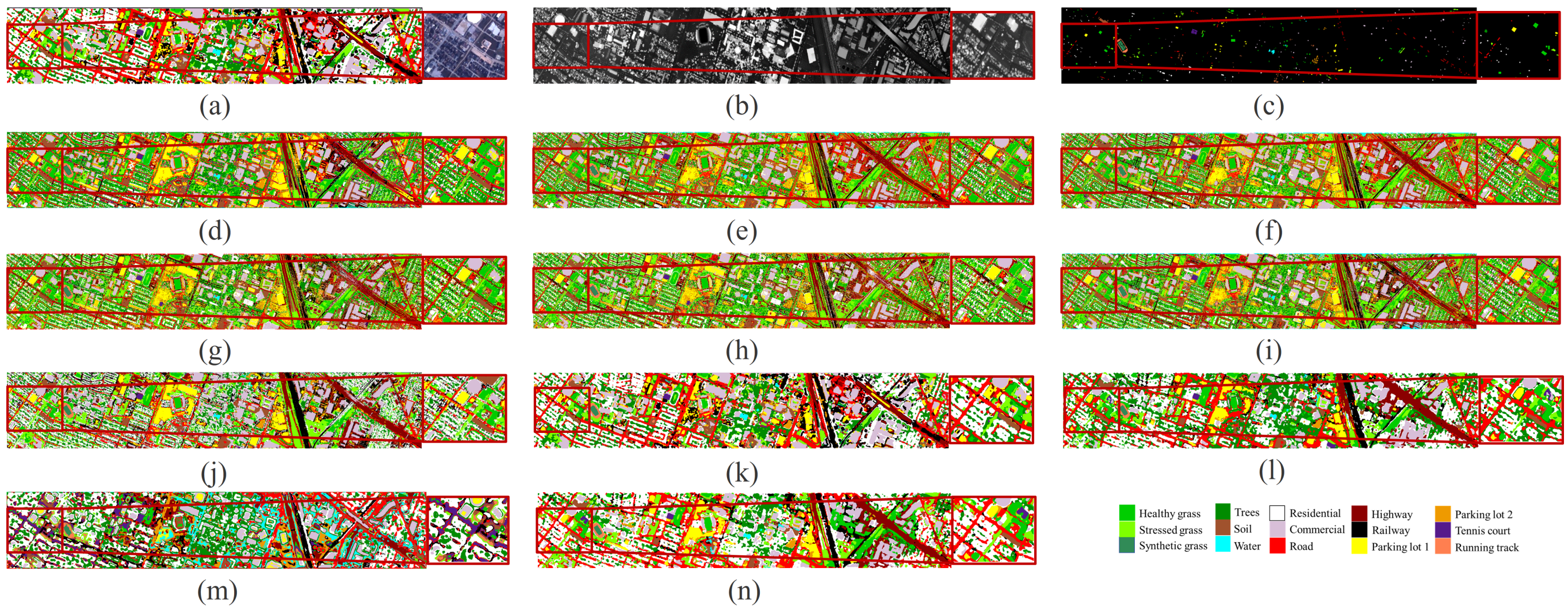

Fig. 6. Classification maps for Houston 2013 dataset. (a) Pseudo-color HSI. (b) LiDAR data. (c) Ground-truth map. (d) GLTNet. (e) EndNet. (f) MDL. (g) S2FL. (h) CoSpace-l1. (i) CoSpace-l2. (j) MSA-GCN. (k) GAMF. (l) HSI. (m) LiDAR. (n) Proposed.

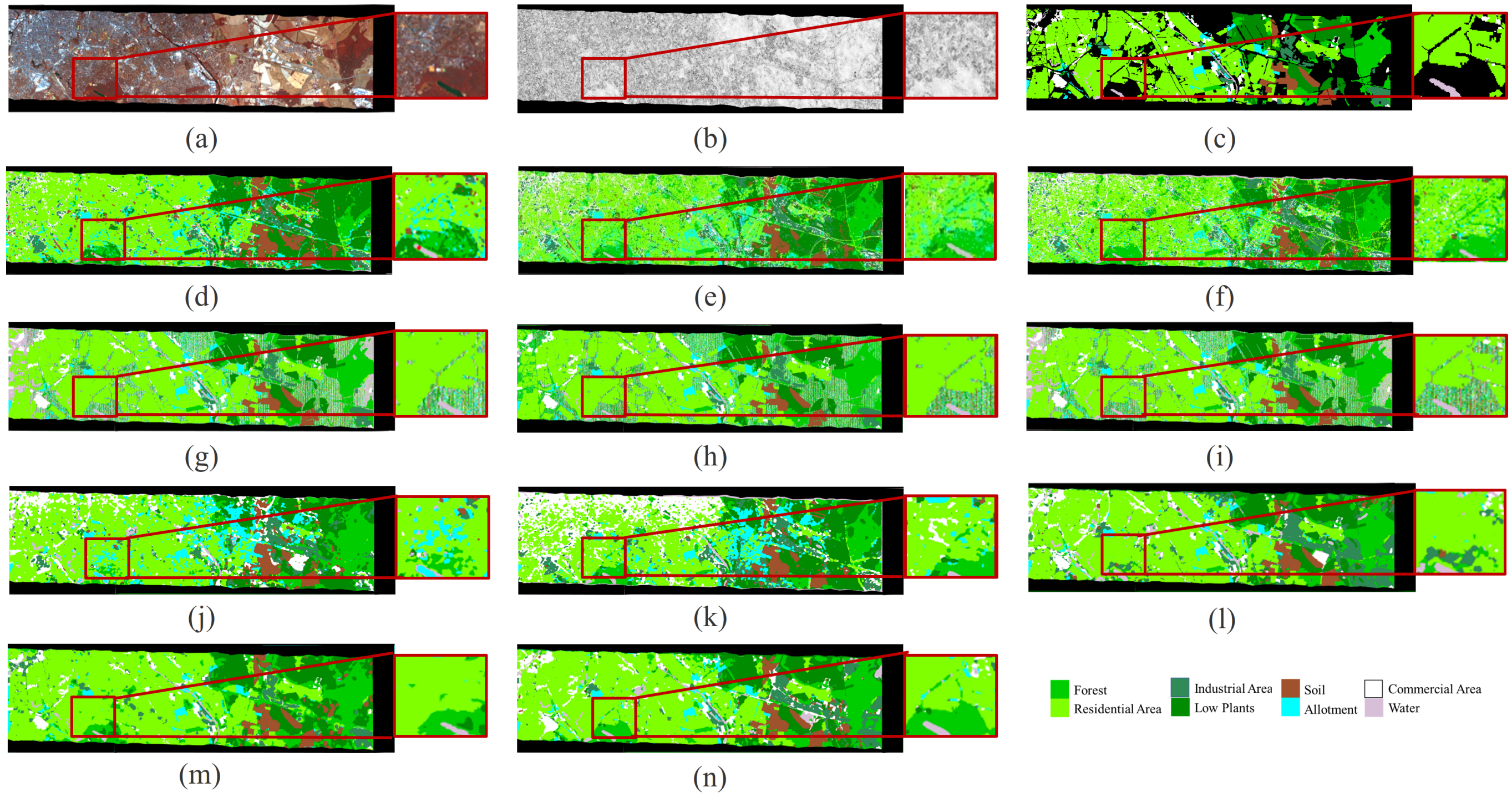

Fig. 7. Classification maps for Berlin dataset. (a) Pseudo-color HSI. (b) SAR data. (c) Ground-truth map. (d) GLTNet. (e) EndNet. (f) MDL. (g) S2FL. (h) CoSpace-l1. (i) CoSpace-l2. (j) MSA-GCN. (k) GAMF. (l) HSI. (m) SAR. (n) Proposed.

### *D. Ablation Study*

The proposed approach consists of three components: the multimodal heterogeneous graph encoder, multi-convolutional modulator, and mean forward mechanism. To demonstrate the performance of each architecture, a series of ablation experiments are performed in three multisource remote sensing datasets in this section.

Table V lists the experimental results of different conditions, in which MHGE is the abbreviation of multimodal heterogeneous graph encoder, MCM represents multi-convolutional modulator, and MF stands for mean forward. It can be noted that the OA and Kappa of the backbone are much lower than in other approaches. When the multimodal heterogeneous graph encoder is added, a significant improvement in OA and Kappa of about 6%-25% and 9%-27% can be gained in all datasets, respectively. This advantage is attributed to the extraction, interaction, and modeling of multimodal discriminative features. With the dynamic salient graph representation, the consistency and complementarity between multiple sources of data can be deeply explored while excluding redundancy. Further improvements in accuracy can be achieved thanks to the addition of the multi-convolutional modulator. Take the difficult disjoint Berlin dataset as an example, the multi-convolutional modulator brings the gain of more than 3% over the non-plus case (i.e., multimodal heterogeneous graph encoder only) for OA and Kappa. The above phenomenon indicates the performance of the multi-convolutional modulator. Owing to the well-designed multi-convolutional structure without self-attention, long-range dependencies can be accurately encoded with less calculation cost. Obviously, the integration of a multimodal heterogeneous graph encoder and multi-convolutional modulator substantially improves OA and Kappa compared to the backbone, for example, it can boost OA by over 9% and Kappa by more than 13% in the Berlin dataset. As expected, the best OA and Kappa are obtained when all components are joined in the whole datasets. The reasons for this effect lie in the fact that the differential feature extraction from multimodal heterogeneous graph encoder, long-distance land cover modeling without heavy computing volume through the multi-convolutional modulator, and lightened overfitting driven by mean forward. The coordinated exploitation of three structures facilitates the breakthrough of the accuracy bottleneck in multimodal land cover identification.

TABLE V
THE EXPERIMENTAL RESULTS OF ABLATION STUDY.

| Method | | Backbone | + MHGE | + MCM | + MF |
|---|---|---|---|---|---|
| MHGE | | | √ | √ | √ |
| MCM | | | | √ | √ |
| MF | | | | | √ |
| Augsburg | OA (%) | 70.65 ±8.32 | 84.45 ±2.87 | 85.14 ±2.59 | **87.39 ±2.36** |
| | Kappa (%) | 60.89 ±11.29 | 78.98 ±3.55 | 79.76 ±3.19 | **82.43 ±3.06** |
| Houston 2013 | OA (%) | 71.52 ±2.86 | 96.08 ±0.76 | 96.72 ±0.59 | **97.08 ±0.49** |
| | Kappa (%) | 69.27 ±3.06 | 95.76 ±0.82 | 96.46 ±0.63 | **96.85 ±0.53** |
| Berlin | OA (%) | 55.65 ±3.00 | 61.83 ±7.56 | 65.25 ±4.12 | **67.75 ±5.58** |
| | Kappa (%) | 37.50 ±2.64 | 47.12 ±5.57 | 50.80 ±4.48 | **53.04 ±4.97** |

Further, the enhancement in classification performance is progressive with the sequential introduction of multimodal heterogeneous graph encoder, multi-convolutional modulator, and mean forward. The OA and Kappa of two constituents are less than those of three architectures but greater than those of one block in three benchmark datasets. The absence of any one of three ingredients will lead to accuracy decay, implying the validity of each construction. Therefore, the all-access to three schemes can accomplish the most profitable outcome for classification.

## E. *Effects of Patch Size and the Number of PCs*

The performance of the model depends to some extent on the choice of parameters, and reasonable parameters can enhance the classification capability. Two essential parameters, patch size and the number of PCs, are investigated with different values in this section.

Fig. 8 depicts the variation of OA with various patch size and number of PCs in all datasets. It can be observed that the OA increases and then decreases as the patch size grows. This transformation can be explicated from two aspects: 1) in the process of increasing the patch size, the model introduces richer spatial information, which makes the OA increase initially. 2) The oversize patch causes more interference and over-smoothing when the patch size keeps enlarging, leading to a reduction in OA instead. Similarly, OA also tends to go up first and then down as the number of PCs rises. This is not surprising because the first few PCs encompass information that is dominant for classification, improving the gain of OA. However, an excessive number of PCs being fed into the network brings the redundancy and noise of the spectral signatures, suppressing the expression of valuable features and thus diminishing OA.

On the basis of the above-mentioned insights, it is crucial to choose the proper patch size and number of PCs for superior classification outcomes. Hence, the patch size is given as 15, 15, and 19 for Augsburg, Houston 2013, and Berlin datasets, respectively. Meanwhile, the number of PCs is set as 32 for all datasets to ensure the characterization ability of the approach.

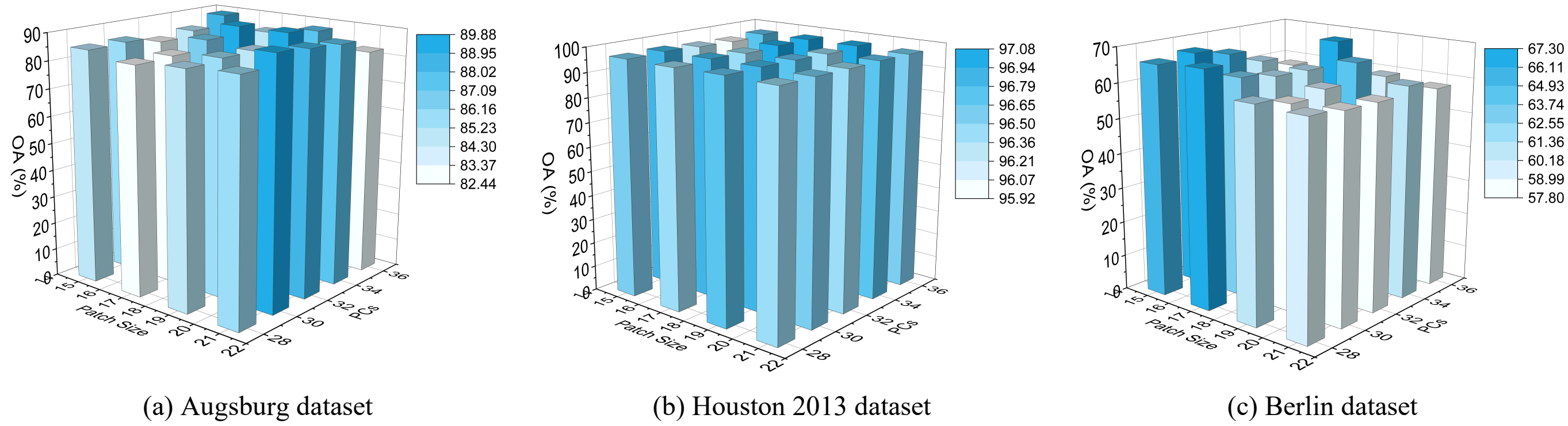

(a) Augsburg dataset (b) Houston 2013 dataset (c) Berlin dataset

Fig. 8. Results of different patch sizes and numbers of PCs.

## F. *Discussion on Patch Size and Efficiency*

To better understand how the proposed THSGR balances performance and computational efficiency, we further conduct an experiment on the patch size and report its impact on both OA and efficiency (FLOPs, MACs, and Params) on the three datasets in Fig. 9.

On the Augsburg dataset, increasing the patch size consistently increases FLOPs and MACs by approximately 2 times from patch size=15 to patch size=21, while OA improves from 83.11% to 88.67%. These results indicate that larger patches can better aggregate local spatial-spectral context in this scene and improve accuracy, but at a clearly higher computational cost. The parameter count is fixed at 3.30 M. For Houston 2013, the trade-off between accuracy and efficiency is more subtle. As the patch size increases from 15 to 19, FLOPs and MACs roughly double, yet OA only slightly improves from 96.23% to a peak of 96.68% at patch size=19. Further enlarging the patch size to 21 continues to increase FLOPs and MACs but slightly decreases OA to 96.45%. This behavior suggests that a moderate patch size is sufficient to capture the dominant spatial-spectral context on Houston 2013. Very small patches under-utilize context, whereas very large patches introduce redundant information that does not bring further gains but still increases computational cost. The parameter count is 5.44 M for various patch sizes. In contrast, Berlin exhibits a different pattern. The smallest patch size=15 already achieves the highest OA of 65.10%, and OA gradually decreases to 58.62% when the patch size is increased to 21, while FLOPs and MACs again approximately double. The parameter count remains constant at 3.57 M. This result suggests that the Berlin scene is more spatially heterogeneous within larger neighborhoods. Using overly large patches tends to mix multiple land-cover types, which can blur boundaries and degrade classification performance while still being less efficient.

Overall, these results verify that the effect of patch size is dataset-dependent and that there is a clear accuracy-efficiency trade-off. Larger patches systematically increase computational cost (FLOPs and MACs), but their impact on OA can be positive, marginal with an optimal medium value, or even negative. This phenomenon further supports the adoption of dataset-specific patch sizes that offer a reasonable compromise between accuracy and efficiency: the patch size is set to 15 for Augsburg and Houston 2013 and to 19 for Berlin.

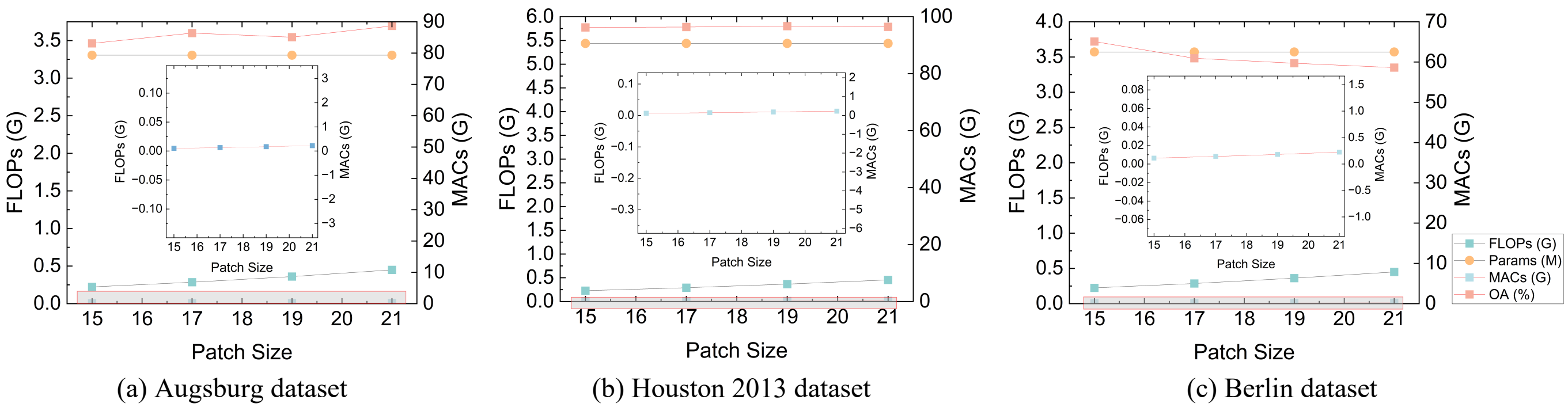

(a) Augsburg dataset (b) Houston 2013 dataset (c) Berlin dataset

Fig. 9. Results of efficiency at different patch sizes.

## G. *Analysis on Accuracy and Efficiency*

To comprehensively evaluate the proposed THSGR, we jointly consider OA and efficiency in terms of FLOPs, MACs, and Params for one sample. Fig. 10 shows these four indicators for all approaches on three datasets.

On the Augsburg dataset, THSGR achieves the highest OA while keeping the computational cost moderate. Specifically, THSGR attains an OA of 87.39%, outperforming the strong GLTNet and MSA-GCN by 6.08% and 10.85%, respectively. Compared with the shallow S2FL, THSGR improves the accuracy by 24.38%. Regarding efficiency, THSGR requires about 0.22 G FLOPs and 3.30 M Params, which is still modest in absolute terms. While this is heavier than GLTNet and MSA-GCN, THSGR remains substantially more efficient than the very deep GAMF: GAMF consumes approximately 3.92 G FLOPs and 9.15 M parameters, whereas THSGR achieves a higher OA with about 18 times fewer FLOPs and 2.7 times fewer Params. This shows that on Augsburg, THSGR delivers a clearly better accuracy-efficiency balance than both traditional shallow approaches and heavy deep architectures. A similar trend can be observed in terms of MACs. The MACs of THSGR remain within the same order of magnitude as those of the compact deep baselines, while being considerably lower than that of the heavy GAMF. This indicates that THSGR achieves its OA gains without incurring a prohibitive increase in arithmetic cost.

On Houston 2013, THSGR again provides the best performance. It reaches 97.09% OA, improving upon GLTNet and MSA-GCN by 2.79% and 1.62%, respectively. Relative to S2FL, THSGR yields a margin of 6.21%. Interestingly, THSGR is also competitive in computational cost on this dataset. It requires only about 0.23 G FLOPs, which is about 5.3 times fewer FLOPs than S2FL. Compared with GAMF, THSGR is both more accurate and more efficient: GAMF reaches 91.06% OA with about 3.09 G FLOPs and 7.31 M parameters, whereas THSGR improves OA by 6.03% and reduces the FLOPs by about 13 times, with slightly fewer parameters. These results indicate that on Houston 2013, THSGR achieves SOTA accuracy with a reduced computational burden relative to the heaviest deep networks. In terms of MACs, THSGR needs about 0.11 G MACs, which is noticeably lower than GAMF (1.54 G MACs).

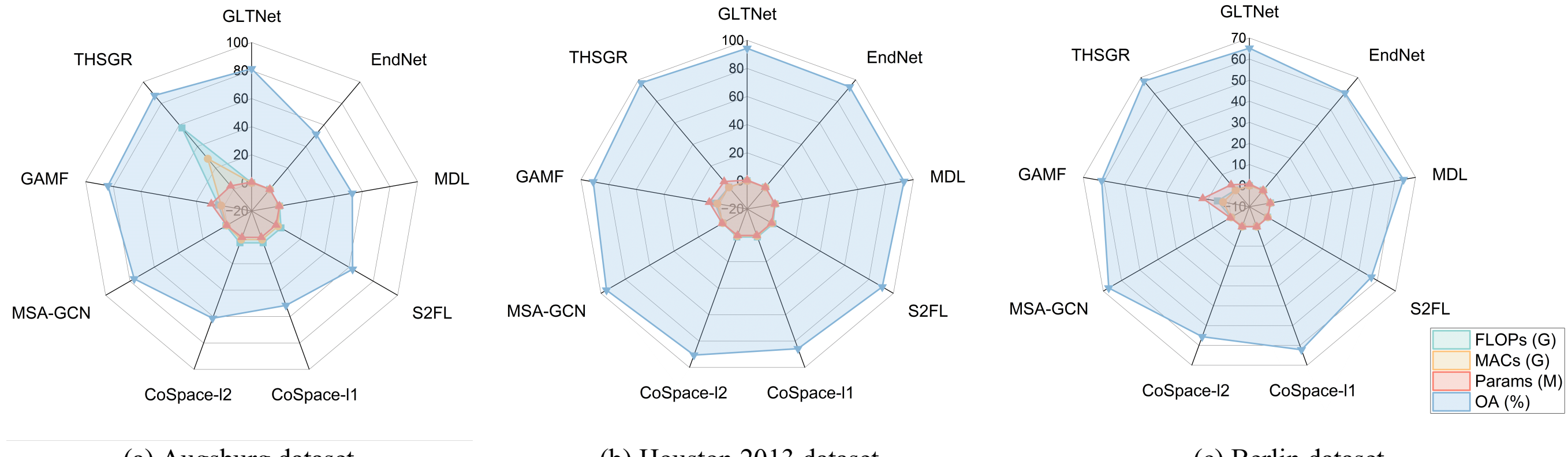

(a) Augsburg dataset (b) Houston 2013 dataset (c) Berlin dataset

Fig. 10. Accuracy and efficiency of diverse algorithms.

On the more challenging disjoint Berlin dataset, THSGR still attains the highest OA. It achieves 67.75% OA, which exceeds GLTNet and MSA-GCN by 2.56% and 0.73%, respectively, and improves over S2FL by 11.00%. This suggests that THSGR is beneficial under the severe domain shift. In terms of complexity, THSGR requires 0.36 G FLOPs and 3.57 M Params. Although this is higher than GLTNet and MSA-GCN, it is still more efficient than the advanced GAMF. Specifically, GAMF consumes 5.41 G FLOPs and 12.43 M Params but only reaches 61.10% OA, whereas THSGR improves OA by 6.75% while using about 15 times fewer FLOPs and 3.4 times fewer Params. This again confirms that THSGR achieves a balance in the accuracy and efficiency. For MACs, THSGR requires 0.36 G MACs, which is about 14.9 times fewer MACs than GAMF while yielding a higher OA.

Across all datasets, THSGR consistently yields the highest OA while maintaining moderate computational cost. Compared with shallow multimodal approaches such as S2FL, CoSpace-l1, and CoSpace-l2, THSGR brings a large accuracy gain with more Params, which is expected. More importantly, relative to heavy deep architectures (e.g., GAMF), THSGR simultaneously improves OA and drastically reduces computational cost, offering a favorable accuracy-efficiency trade-off.

### *H. Discussions about Complexity between MSA and Multi-Convolutional Modulator*

To evaluate the complexity of the popular MSA and the proposed multi-convolutional modulator, the experiments in three datasets concerning FLOPs and Params are discussed in this section.

Fig. 11 visually illustrates the discrepancy between MSA and the presented multi-convolutional modulator. The larger the pattern size, the larger the value. Green, purple, and orange are assigned to distinguish Augsburg, Houston 2013, and Berlin datasets. As anticipated, the FLOPs and Params of the proposed structure are much smaller than MSA in all datasets. This occurrence is credited to the local connectivity and weight sharing of the designed approach. On the one hand, a multi-convolutional modulator performs multiplication and addition operations only in the local receptive field. Instead, MSA proceeds one by one for all input data, which produces a large number of redundant computations and a dramatic increase in FLOPs. Besides, unnecessary calculations also introduce interference or noise, which degrades the accuracy. On the other hand, all elements on the one feature map adopt the same convolution kernel in a multi-convolutional modulator, i.e., weight sharing. That means the number of parameters can be greatly reduced. By contrast, MSA computes parameters based on the entire input data, and the high-dimensional HSI as input further burdens the number of parameters. In this case, the Params of MSA is greater than that of the multi-convolutional modulator, which limits the applicability. Also, these experimental results also empirically verify the previous theoretical complexity comparison.

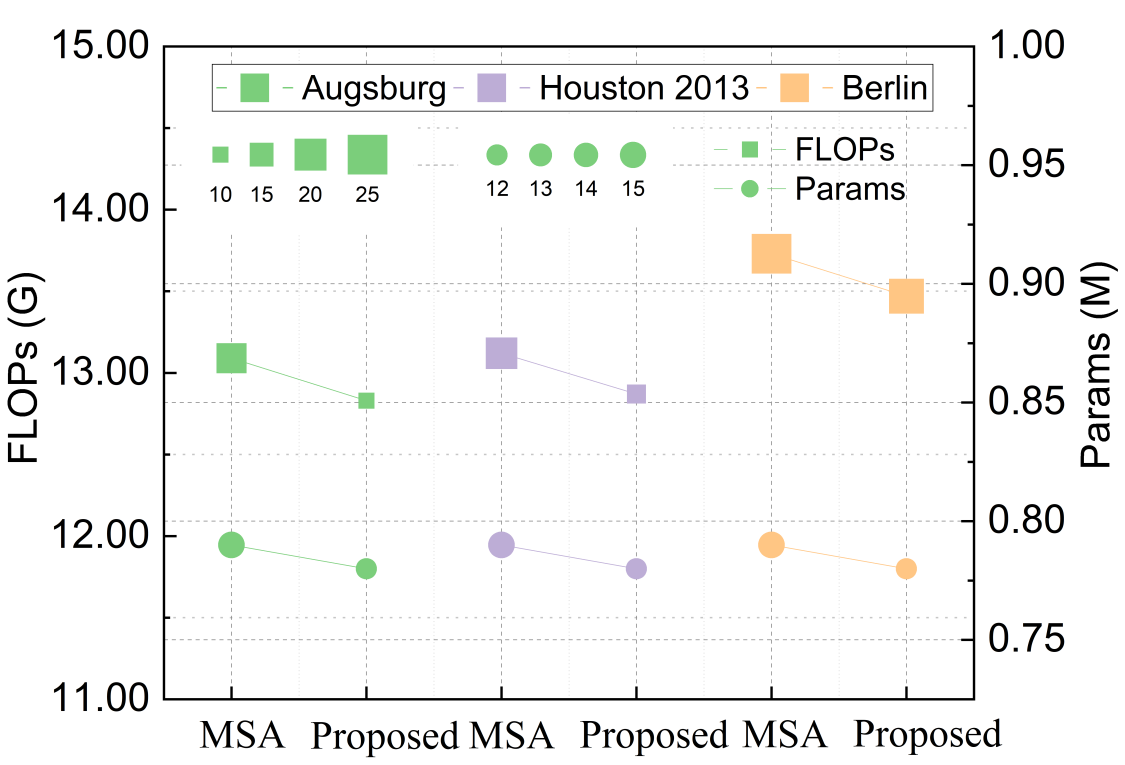

Fig. 11. The comparisons of FLOPs and Params between MSA and multi-convolutional modulator.

According to the aforesaid discussions, the elaborate multi-convolutional modulator can reduce FLOPs and Params, alleviating complexity without more accuracy loss.

## IV. Conclusion

In the field of HSI and SAR/LiDAR joint classification, there are still the following problems that hinder the performance gain: 1) non-discriminatory representation of heterogeneous features, 2) redundant calculations and features when modeling long-term dependencies, and 3) overfitting stemming from limited labeled samples. To address the above three issues, the multimodal heterogeneous graph encoder, self-attention-free multi-convolutional modulator, and mean forward mechanism are proposed in correspondence. Based on these structures, THSGR is built to cross the modal gap and model intrinsic consistency and complementarity. In a simple and effective fashion, the distinctive features of multi-source data, distant but geographically tightly related ground targets, and the mitigation of over-smoothing can be simultaneously gained for better land cover identification. Extensive experiments and analyses in three benchmark datasets can verify the performance of the proposed approach. For future work, we expect to incorporate more modalities to research weakly supervised multi-source remote sensing image classification methods.